\documentclass{l4dc2022}
\usepackage{times, multirow}

\title[Continuous-Time Policy Gradient for Optimisation of Structured Neural Controller]{Optimisation of Structured Neural Controller Based on Continuous-Time Policy Gradient}

\author{
 \Name{Namhoon Cho} \Email{n.cho@cranfield.ac.uk} \\ \Name{Hyo-Sang Shin} \Email{h.shin@cranfield.ac.uk}\\
 \addr Centre for Autonomous and Cyber-Physical Systems, School of Aerospace, Transport and Manufacturing, Cranfield University, Cranfield, Bedfordshire, MK43 0AL, United Kingdom
}

\begin{document}
\maketitle

\begin{abstract}
This study\footnote{This work has been submitted to the IEEE for possible publication. Copyright may be transferred without notice, after which this version may no longer be accessible.} presents a policy optimisation framework for structured nonlinear control of continuous-time (deterministic) dynamic systems. 
The proposed approach prescribes a structure for the controller based on relevant scientific knowledge (such as Lyapunov stability theory or domain experiences) while considering the tunable elements inside the given structure as the point of parametrisation with neural networks. To optimise a cost represented as a function of the neural network weights, the proposed approach utilises the continuous-time policy gradient method based on adjoint sensitivity analysis as a means for correct and performant computation of cost gradient. 
This enables combining the stability, robustness, and physical interpretability of an analytically-derived structure for the feedback controller with the representational flexibility and optimised resulting performance provided by machine learning techniques. 
Such a hybrid paradigm for fixed-structure control synthesis is particularly useful for optimising adaptive nonlinear controllers to achieve improved performance in online operation, an area where the existing theory prevails the design of structure while lacking clear analytical understandings about tuning of the gains and the uncertainty model basis functions that govern the performance characteristics. Numerical experiments on aerospace applications illustrate the utility of the structured nonlinear controller optimisation framework.
\end{abstract}


\section{Introduction} \label{Sec:Intro}
Recent technological advances that realise differentiable programming paradigm have led to the resurgence of optimise-then-discretise approaches towards solving learning/control problems associated with continuous-time dynamic systems described by Ordinary Differential Equations (ODEs). On one hand, the philosophy of combining structured scientific models given by differential equations with unstructured data-centric machine learning models has been appreciated for its effectiveness in long-term predictive accuracy and learning efficiency \cite{Rackauckas_2021}. On the other hand, methods exploiting the characteristics of continuous-time dynamic systems are shown to be promising in learning optimal control for physical systems that evolve in continuous time in comparison to those dependent upon explicit time-discretisation \cite{Ainsworth_2021,Kim_2020,Kim_2021,Doya_2000, Wang_2020,Yildiz_2021,Lee_2021}, most likely because the loss/cost gradient computation is more accurate and physically consistent. Modern machine learning tools such as \texttt{DiffEqFlux.jl} implemented in \texttt{Julia} support these two directions -- i) scientific machine learning approach applied to ii) the Continuous-Time Policy Gradient (CTPG) method -- through combination of Algorithmic Differentiation (AD) and adaptive time-stepping ODE solver \cite{Innes_2018,Innes_2019a,Innes_2019b,Ma_2021}. AD technologies composed with neighbouring software packages for solving ODEs, constructing Neural Networks (NN), and optimisation enable to address problems through abstract description in a high-level language.

With this background, a promising direction for feedback controller synthesis is to formulate a learning problem by closing the loop with NN on the signal path and to leverage the optimality condition described in the most relevant form respecting the system continuity properties for training. The online learning \cite{Wagener_2019} or inference \cite{Todorov_2008, Levine_2018} perspective on control has been well-established based on the unified formal notion of optimisation. Also, the cost gradient evaluation procedure involving backpropagation through the Pontryagin's minimum principle has been studied in \cite{Jin_2020, Jin_2021}. However, these earlier studies considered discrete-time setting, which might be rendered ineffective in complex physical systems if fine discretisation is necessary or when the effect of the interval size is not clear. The CTPG method explored in \cite{Ainsworth_2021, Sandoval_2021} aims to overcome the difficulties of discretise-then-optimise style by performing differentiation through physics-based ODE model involving NN components for computing the cost gradient with respect to policy parameters with the aid of AD. More specifically, the CTPG approach adopts the adjoint sensitivity analysis technique also known as the Kelley-Bryson method \cite{Dreyfus_1990} by using adaptive step ODE solvers for forward/backward passes. This approach avoids errors and resolution dependence due to arbitrary discretisation in dealing with dynamic systems that are continuous in time, state, and input variables. Hence, the CTPG method allows for simplified and accelerated optimisation of a parametric policy of certain classes. 
Interestingly, a technique developed for nonlinear suboptimal control of aerospace vehicles solves the ODE for output sensitivity matrix backward similarly as done in the CTPG method, but only performs incremental update of the open-loop form profile for the control input instead of training any NN policy \cite{Maity_2014}.

This study aims to further strengthen the capability of policy learning approach based on the CTPG method by prescribing the controller structure and parametrising the tunable elements with NN. The basic premise of the structured controller optimisation is that incorporating the domain knowledge of control theory provides a nice search space leading to many benefits in performance and robustness during operation as well as in improved efficiency of learning process, as empirically evidenced in earlier studies such as \cite{Roberts_2011, Shin_2020, KimSH_2020,Wang_2021}. The expected advantages of the CTPG-based fixed-structure design approach are threefold. From the stability perspective, specifying a controller structure with the one that guarantees closed-loop stability can maintain the certifiability of resulting controller, reduce the sensitivity of response characteristics with respect to NN weights, and reduce the number of failure cases in the initial learning stage. From the viewpoint of practical applications, practitioners may find the structured policy more physically interpretable and hence reliable while the tunable elements given as smooth functions of operating condition are optimised against a given performance objective. With regard to the tuning difficulties, the proposed approach circumvents the tedious and time-consuming process of parameter scheduling for controllers of known structures in the absence of clear analytical criteria, the situation usually encountered in adaptive nonlinear control. Furthermore, the proposed approach complements the method of \cite{Sanchez-Sanchez_2018,Izzo_2021,Gaudet_2020} to learn optimal controllers by fitting NN to the separately-generated optimal control database for which the end result might inevitably lose optimality reasonings due to approximation inaccuracies in the later stage.

The rest of this paper is organised as follows: Section \ref{Sec:CTPG} briefly reviews the derivation of the CTPG method which is essentially the adjoint state method for (local) sensitivity analysis. The description particularly emphasises the modern soft-optimality formulation that includes a NN weight-dependent regularisor alongside the traditional Bolza form cost functional. Section \ref{Sec:StructOC_CTPG} develops a structured nonlinear controller optimisation framework based on the CTPG method by parametrising the tunable elements in a prescribed controller structure with NN instead of the actual control input. Numerical experiments in Sec. \ref{Sec:NumSim} demonstrates the effectiveness of the proposed approach in the context of aerospace applications.

\section{Continuous-Time Policy Gradient} \label{Sec:CTPG}
When control problems are approached through the lens of learning formalism while adopting NN in the loop, the optimal control cost function that represents the desired performance objective in terms of the state and input variables becomes the loss function for NN training in terms of the weights. Given the gradient of the loss function with respect to the NN parameters at each iteration, a generic gradient-based algorithm can be employed to optimise the parameters. Hence, correct and efficient evaluation of the loss gradient is imperative.

The gradient of the scalar loss function can be obtained by leveraging the adjoint state method which originates from the same mathematical foundation with the Pontryagin minimum principle for optimal control \cite{Dreyfus_1990,Cao_2003}. This reverse mode technique circumvents the necessity to compute the solution sensitivity with respect to each parameter, the procedure which becomes rapidly computationally expensive in the forward mode technique as the number of parameters increases. The Kelley-Bryson theory of the adjoint state method in conjunction with the recently developed differentiable programming systems establishes the CTPG method suitable for modern NN applications involving a large number of parameters \cite{Ainsworth_2021}. This desideratum is achieved by combining NN and ODE with adaptive step solvers while maintaining the transparence of the entire processing pipeline for AD \cite{Innes_2019b}.

Consider the following optimisation problem:
\begin{equation} \label{Eq:Prob_CTPG}
	\begin{aligned}
		\underset{\mathbf{p}}{\text{minimise}} && J\left(\mathbf{p}\right) &= \phi\left(\mathbf{x}\left(t_{f}\right)\right) + \int_{t_{0}}^{t_{f}} L\left(\tau, \mathbf{x}\left(\tau\right), \mathbf{p}\right) d\tau + R\left(\mathbf{p}\right)\\
		\text{subject to} && \dot{\mathbf{x}}\left(t\right) &= \mathbf{f}\left( t, \mathbf{x}\left(t\right), \mathbf{p} \right)\\
		&& \mathbf{x}\left(t_{0}\right) &= \mathbf{x}_{0} \text{~: fixed}\\
		&& & t_{0}, ~t_{f} \text{~: fixed}
	\end{aligned}
\end{equation}
where the variables $t$, $\mathbf{x}$, and $\mathbf{p}$ denote the time, the state, and the parameter, respectively, and the functions $J$, $\phi$, $L$, $R$, and $\mathbf{f}$ represent the total cost, the terminal cost, the running cost, the regularisor, and the system dynamics, respectively. The purpose of adjoint sensitivity analysis is to evaluate the gradient $\nabla_{\mathbf{p}}J := \frac{d J}{d \mathbf{p}}$. 

The method of Lagrange multiplier in the function space leads to the augmented cost function
\begin{equation} \label{Eq:aug_cost}
	\resizebox{0.94\hsize}{!}{$\displaystyle
	\begin{aligned}
	J_{aug}\left(\mathbf{p}\right) &= J\left(\mathbf{p}\right) - \int_{t_{0}}^{t_{f}}\boldsymbol{\lambda}^{T}\left(\tau\right)\left\{\dot{\mathbf{x}}\left(\tau\right) - \mathbf{f}\left( \tau, \mathbf{x}\left(\tau\right), \mathbf{p} \right)\right\}d\tau\\
	&= \phi\left(\mathbf{x}\left(t_{f}\right)\right) + R\left(\mathbf{p}\right) + \int_{t_{0}}^{t_{f}} \left[L\left(\tau, \mathbf{x}\left(\tau\right), \mathbf{p}\right) - {\boldsymbol{\lambda}}^{T}\left(\tau\right)\left\{\dot{\mathbf{x}}\left(\tau\right) - \mathbf{f}\left( \tau, \mathbf{x}\left(\tau\right), \mathbf{p} \right)\right\}\right] d\tau 
	\end{aligned}
	$}
\end{equation}
where $\boldsymbol{\lambda}$ denotes the costate. Considering the relation $J_{aug} = J$, the Leibniz rule for differentiation under the integral sign along with the chain rule of differentiation yields
\begin{equation} \label{Eq:grad_aug_cost}
	\frac{dJ}{d\mathbf{p}} = \frac{d J_{aug}}{d \mathbf{p}} = \frac{\partial \phi}{\partial \mathbf{x}\left(t_{f}\right)}\mathbf{S}\left(t_{f}\right) + \frac{dR}{d\mathbf{p}} + \int_{t_{0}}^{t_{f}} \left\{ \frac{\partial L}{\partial \mathbf{x}} \mathbf{S} + \frac{\partial L}{\partial \mathbf{p}} - \boldsymbol{\lambda}^{T}\left(\dot{\mathbf{S}} - \frac{\partial \mathbf{f}}{\partial \mathbf{x}}\mathbf{S} - \frac{\partial \mathbf{f}}{\partial \mathbf{p}} \right) \right\}d\tau
\end{equation}
where $\mathbf{S}\left(t\right) := \frac{d \mathbf{x}\left(t\right)}{d \mathbf{p}}$ represents the solution sensitivity. By applying integration by parts, we have
\begin{equation} \label{Eq:int_parts}
	\int_{t_{0}}^{t_{f}} \boldsymbol{\lambda}^{T}\dot{\mathbf{S}}d\tau = \left.\boldsymbol{\lambda}^{T}\mathbf{S}\right|_{t_{0}}^{t_{f}} - \int_{t_{0}}^{t_{f}} \dot{\boldsymbol{\lambda}}^{T}\mathbf{S}d\tau
\end{equation}
Substituting Eq. (\ref{Eq:int_parts}) into Eq. (\ref{Eq:grad_aug_cost}) while considering the fact that $\mathbf{S}\left(t_{0}\right) = \mathbf{0}$ due to the fixed initial condition $\mathbf{x}_{0}$ leads to
\begin{equation} \label{Eq:grad_aug_cost_with_int_parts}
	\begin{aligned}
	\frac{dJ}{d\mathbf{p}} &= \left(\frac{\partial \phi}{\partial \mathbf{x}\left(t_{f}\right)} -\boldsymbol{\lambda}^{T}\left(t_{f}\right) \right)\mathbf{S}\left(t_{f}\right) + \frac{dR}{d\mathbf{p}} \\
	&\quad + \int_{t_{0}}^{t_{f}} \left\{  \frac{\partial L}{\partial \mathbf{p}} + \boldsymbol{\lambda}^{T}\frac{\partial \mathbf{f}}{\partial \mathbf{p}} + \left(\frac{\partial L}{\partial \mathbf{x}} + \boldsymbol{\lambda}^{T}\frac{\partial \mathbf{f}}{\partial \mathbf{x}} + \dot{\boldsymbol{\lambda}}^{T} \right)\mathbf{S} \right\}d\tau
	\end{aligned}
\end{equation}
To remove any dependence to the solution sensitivity from Eq. (\ref{Eq:grad_aug_cost_with_int_parts}), one can choose $\boldsymbol{\lambda}$ to satisfy 
\begin{equation} \label{Eq:dlambda_dt}
	\begin{aligned}
		\dot{\boldsymbol{\lambda}}\left(t\right) &= -\left[\frac{\partial \mathbf{f}}{\partial \mathbf{x}}\right]^{T}\boldsymbol{\lambda}\left(t\right) - \left[\frac{\partial L}{\partial \mathbf{x}}\right]^{T}\\
		\boldsymbol{\lambda}\left(t_{f}\right) &= \left[\frac{\partial \phi}{\partial \mathbf{x}\left(t_{f}\right)} \right]^{T}
	\end{aligned}
\end{equation}
Consequently, the gradient is given by
\begin{equation} \label{Eq:dJ_dp}
	\frac{dJ}{d\mathbf{p}} = \frac{dR}{d\mathbf{p}} + \int_{t_{0}}^{t_{f}} \left(  \frac{\partial L}{\partial \mathbf{p}} + \boldsymbol{\lambda}^{T}\frac{\partial \mathbf{f}}{\partial \mathbf{p}} \right)d\tau
\end{equation}

The backward pass of solving the costate dynamics in Eq. (\ref{Eq:dlambda_dt}) from the final boundary condition requires the state trajectory for evaluation of the partial derivatives. A preceding forward pass suffices the purpose (See Remark \ref{Rem:AdjImpl}). One may resort to either algorithmic or analytic differentiation for constructing the partial derivative evaluation code. Algorithm \ref{Algo:Reg_CTPG} summarises the forward-backward procedure for the CTPG with regularisation.
\begin{algorithm2e}[ht!]
	\caption{Regularised Continuous-Time Policy Gradient ($\mathsf{R}$-$\mathsf{CTPG}$)}
	\label{Algo:Reg_CTPG}
	\DontPrintSemicolon
	\LinesNumbered

	\KwIn{$\mathbf{p}$}
	\KwData{$\phi$, $L$, $R$, $\mathbf{f}$, $t_{0}$, $t_{f}$, $\mathbf{x}_{0}$, $\mathsf{ODE\_solver}\left(\mathsf{ODE}, \mathsf{IC}, t_{\text{span}}  \right)$}
	
	\KwOut{$J$, $\frac{dJ}{d\mathbf{p}}$}
	\BlankLine
	
	\Begin(\textsf{Forward Pass}){
		$\mathbf{f}_{\mathbf{z}} \left( t, \mathbf{x}\left(t\right), \mathbf{p} \right) := \begin{bmatrix}
			\mathbf{f}\left( t, \mathbf{x}\left(t\right), \mathbf{p} \right)\\
			L\left( t, \mathbf{x}\left(t\right), \mathbf{p} \right)
		\end{bmatrix}$, \quad $\mathbf{z}_{0} \leftarrow \begin{bmatrix}
			\mathbf{x}_{0}\\
			0
		\end{bmatrix}$\;
		$\mathbf{z}\left(t \in \left[t_{0}, t_{f}\right]\right) \leftarrow \mathsf{ODE\_solver}\left(\mathbf{f}_{\mathbf{z}}\left(t, \mathbf{x}\left(t\right),\mathbf{p}\right), \mathbf{z}_{0}, \left[t_{0}, t_{f}\right] \right)$\;
		$\mathbf{x}\left(t\in \left[t_{0}, t_{f}\right]\right) \leftarrow \mathbf{z}\left(t\in \left[t_{0}, t_{f}\right]\right)\left[1:\dim\left(\mathbf{x}\right)\right]$\;
		$J \leftarrow \phi\left(\mathbf{x}\left(t_{f}\right)\right) + R\left(\mathbf{p} \right) + \mathbf{z}\left(t_{f}\right)\left[\dim\left(\mathbf{x}\right)+1\right]$\;
	}
	
	\Begin(\textsf{Backward Pass}){
		$\mathbf{f}_{\mathbf{w}} \left( t, \mathbf{x}\left(t\right), \boldsymbol{\lambda}\left(t\right), \mathbf{p} \right) := \begin{bmatrix}
			-\left[\frac{\partial \mathbf{f}}{\partial \mathbf{x}}\right]^{T}\boldsymbol{\lambda}\left(t\right) - \left[\frac{\partial L}{\partial \mathbf{x}}\right]^{T}\\[0.3em]
			- \left[\frac{\partial \mathbf{f}}{\partial \mathbf{p}}\right]^{T}\boldsymbol{\lambda}\left(t\right) - \left[\frac{\partial L}{\partial \mathbf{p}}\right]^{T} 
		\end{bmatrix}$, \quad $\mathbf{w}_{f} \leftarrow \begin{bmatrix}
		\left[\frac{\partial \phi}{\partial \mathbf{x}\left(t_{f}\right)} \right]^{T}\\
		\mathbf{0}
		\end{bmatrix}$\;
		$\mathbf{w}\left(t_{0}\right) \leftarrow \mathsf{ODE\_solver}\left(\mathbf{f}_{\mathbf{w}}\left(t, \mathbf{x}\left(t\right), \boldsymbol{\lambda}\left(t\right), \mathbf{p}\right), \mathbf{w}_{f}, \left[t_{f}, t_{0}\right]\right)$\;
		$\left[\frac{dJ}{d\mathbf{p}}\right]^{T} \leftarrow \left[\frac{dR}{d\mathbf{p}}\right]^{T} + \mathbf{w}\left(t_{0}\right)\left[\dim\left(\mathbf{x}\right) +1 : \dim\left(\mathbf{x}\right)+\dim\left(\mathbf{p}\right) \right]$\;
	}

	\Return{$J\left(\mathbf{p}\right)$, $\frac{dJ}{d\mathbf{p}}\left(\mathbf{p}\right)$}\;
\end{algorithm2e}

\section{Structured Policy Optimisation with Continuous-Time Policy Gradients} \label{Sec:StructOC_CTPG}
Originally, an optimal control problem aims to find a control input function $\mathbf{u}\left(t\right)$ that minimises a cost functional subject to the system dynamics. One approximate optimal control approach is to convert a function space optimisation problem into one in parameter space by introducing parametric representation for the objects comprising the problem description. The possible ways of control input parametrisation can be classified as follows:
\begin{itemize}
	\item open-loop form control input: $\mathbf{u}\left(t\right) = \pi\left(t,\mathbf{p}\right)$
	\item closed-loop form control input: $\mathbf{u}\left(t\right) = \pi\left(\mathbf{x}\left(t\right),\mathbf{p}\right)$
	\item controller tunable elements: $\mathbf{u}\left(t\right) = \mathcal{C}\left(\mathbf{K}\left(t\right), \mathbf{x}\left(t\right)\right)$ with $\mathbf{K}\left(t\right) = \pi\left(t, \mathbf{x}\left(t\right), \mathbf{p}\right)$
\end{itemize}

This study considers direct optimisation of neural policies for tunable elements in a prescribed structure controller. Adopting a moderate-size NN in the control loop may confine the best achievable performance to the extent depending on the functional class of the NN representation. That is, the training result might still be a suboptimal solution as compared with the original exact optimal control problem. Nevertheless, directly optimising a feedback policy is considered more useful for actual practice than obtaining the optimal control solution in the form of time-indexed profiles; i) incorporating feedback renders the closed-loop system more robust against uncertainties, thus facilitating deployment, and ii) a separate controller for reference trajectory tracking is not necessary. 

\begin{figure}[ht!]
	\begin{center}
		\includegraphics[width = 0.4\textwidth]{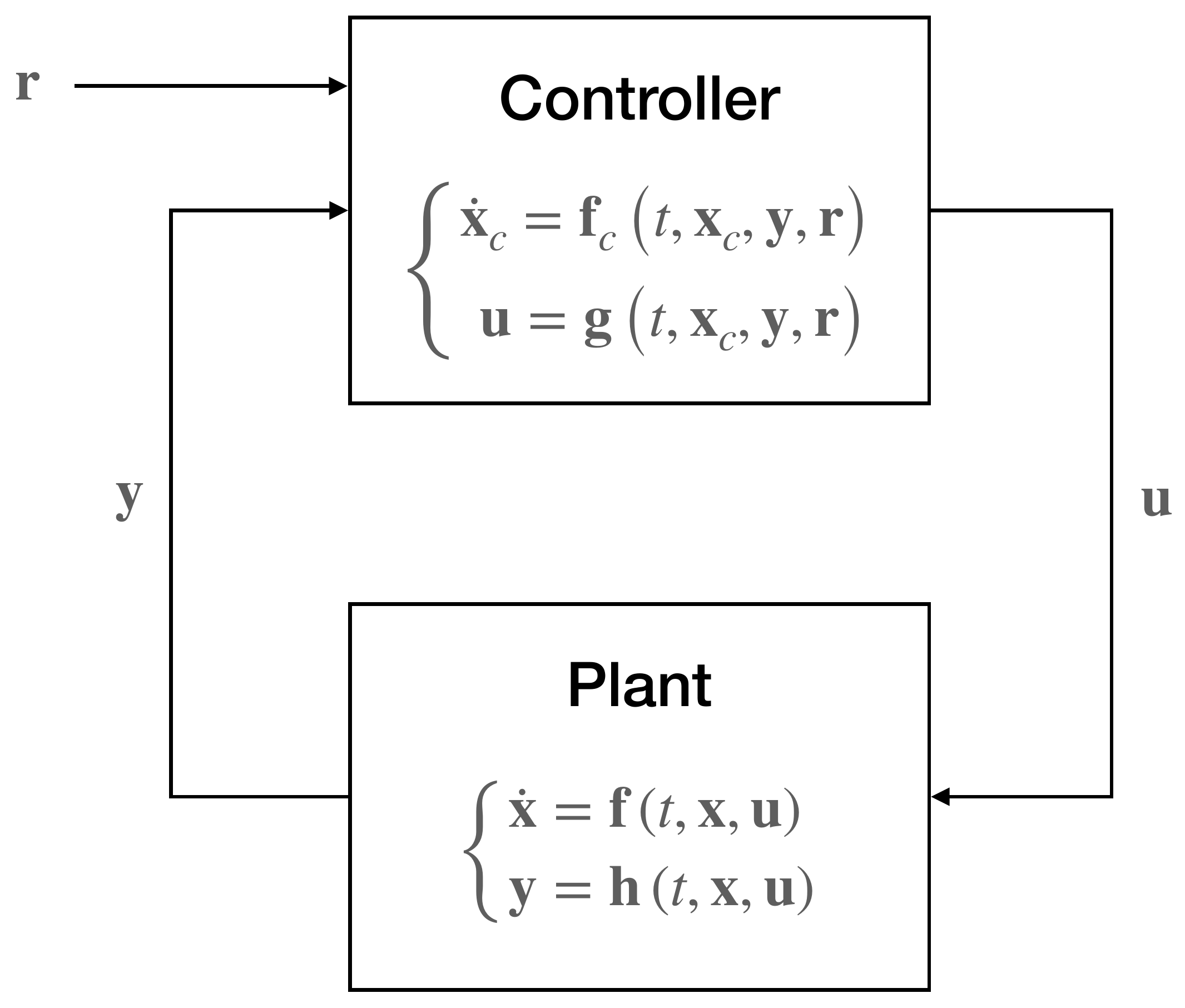}
		\caption{Block Diagram of Closed-Loop System}
		\label{Fig_BlkDiag_CL}
	\end{center}
\end{figure}

Consider the closed-loop system given by feedback interconnection of a plant and a controller as depicted in Fig. \ref{Fig_BlkDiag_CL}. A structured controller optimisation problem can generally be formulated as follows
\begin{equation} \label{Eq:Prob_StructOC}
	\begin{aligned}
		\underset{\mathbf{p}}{\text{minimise}} && J\left(\mathbf{p}\right) &= \mathbb{E}_{\mathbf{x}_{0},\, \mathbf{r}} \left[\phi\left(\mathbf{x}\left(t_{f}\right)\right) + \int_{t_{0}}^{t_{f}} L\left(\tau, \mathbf{x}\left(\tau\right), \mathbf{u}\left(\tau\right)\right) d\tau\right] + R\left(\mathbf{p}\right)\\
		\text{subject to} && \dot{\mathbf{x}}\left(t\right) &= \mathbf{f}\left( t, \mathbf{x}\left(t\right), \mathbf{u}\left(t\right) \right)\\
		&& \mathbf{y}\left(t\right) &= \mathbf{h}\left(t,\mathbf{x}\left(t\right), \mathbf{u}\left(t\right)\right)\\
		&& \dot{\mathbf{x}}_{c}\left(t\right) &= \mathbf{f}_{c}\left(t, \mathbf{x}_{c}\left(t\right), \mathbf{y}\left(t\right), \mathbf{r}\left(t\right)\right),\quad \mathbf{r}\left(t\right) \in \mathcal{R}\\
		&& \mathbf{u}\left(t\right) &= \mathbf{g}\left(t, \mathbf{x}_{c}\left(t\right), \mathbf{y}\left(t\right), \mathbf{r}\left(t\right)\right),\quad \left(\mathbf{f}_{c}, \mathbf{g}\right) \in \mathcal{K}\left(\pi\left(\mathbf{p}\right)\right)\\
		&& \mathbf{x}\left(t_{0}\right) &= \mathbf{x}_{0} \in \mathcal{X}_{0}, \quad \mathbf{x}_{c}\left(t_{0}\right) = \mathbf{x}_{c_{0}} \text{~: fixed}\\
		&& & t_{0}, ~t_{f}~ \text{: fixed}
	\end{aligned}
\end{equation}
where $\mathbf{x}$, $\mathbf{y}$, $\mathbf{u}$, $\mathbf{x}_{c}$, and $\mathbf{r}$ denote the state, the output, the input, the controller state, and the exogenous reference, respectively. 

In Eq. (\ref{Eq:Prob_StructOC}), $\mathcal{K}$ represents the set of functions having a designer-defined structure and $\pi$ denotes the NN component. For example, if one desires to develop a Linear Parameter-Varying (LPV) controller, the policy class can be defined as
\begin{equation} \label{Eq:LPV_structure}
	\mathcal{K}_{LPV} ~:~ 
	\begin{bmatrix}
		\mathbf{f}_{c}\\
		\mathbf{g}
	\end{bmatrix}
	 = \pi\left(\mathbf{y}\left(t\right), \mathbf{p}\right)
	\begin{bmatrix}
		\mathbf{x}_{c}\left(t\right)\\
		\mathbf{y}\left(t\right)\\
		\mathbf{r}\left(t\right)
	\end{bmatrix}
\end{equation}
with an appropriate dimension for the matrix $\pi$ which corresponds to the controller gains. Note that optimisation of the NN parameters with a loss function that promotes sparsity of the output can potentially be useful to find a good loop structure for the given control problem within the prescribed class of fixed-order linear controllers. In the case of composite adaptive control for a dynamic system model containing additive linearly-parametrised uncertainty $\Delta\left(\mathbf{x}\left(t\right)\right) = \mathbf{\Phi}^{T}\left(\mathbf{x}\left(t\right)\right)\boldsymbol{\theta}$ in the input channel, one possible form for the structured controller parametrisation is
\begin{equation} \label{Eq:CAdC_structure}
	\mathcal{K}_{CAd} ~:~ 
	\begin{bmatrix}
		\mathbf{f}_{c}\\
		\mathbf{g}
	\end{bmatrix}
	 = 
	\begin{bmatrix}
		\mathbf{m}_{\text{direct}}\left(\cdot\right) - G_{LPF}\left[s; \pi\left(t,\mathbf{x}_{c},\mathbf{y}, \mathbf{p}\right)\right] \mathbf{\Phi}\left(\mathbf{x}\right) \left\{ \mathbf{\Phi}^{T}\left(\mathbf{x}\right)\mathbf{x}_{c}\left(t\right) - \mathcal{Y}\left(t\right)\right\}\\[0.3em]
		\mathbf{u}_{\text{base}}\left(\cdot\right) - \mathbf{\Phi}^{T}\left(\mathbf{x}\right)\mathbf{x}_{c}\left(t\right)
	\end{bmatrix}
\end{equation}
where $\mathbf{u}_{\text{base}}\left(\cdot\right)$ is a baseline controller that stabilises the nominal system, $\mathbf{m}_{\text{direct}}\left(\cdot\right)$ is a direct adaptation term that is defined appropriately to ensure asymptotic tracking error convergence, and $\mathcal{Y}\left(t\right)$ is an observation for the uncertainty $\Delta\left(\mathbf{x}\left(t\right)\right)$. The neural tuning element resides in the low-pass filter denoted by $G_{LPF}$.

Note that the cost function $J$ in Eq. (\ref{Eq:Prob_StructOC}) is given as a sample mean over various combinations of $\left(\mathbf{x}_{0}, \mathbf{r}\left(t\right)\right)$ to consider multiple operating conditions. Here, $\mathcal{X}_{0}$ and $\mathcal{R}$ refer to the set of initial conditions and reference signals in the operational range, respectively, which can either be a deterministic collection over a grid or samples of stochastic distribution.

The problem in Eq. (\ref{Eq:Prob_StructOC}) can be solved with a gradient-descent-based optimiser such as ADAM or L-BFGS by using the CTPG method described in Algorithm \ref{Algo:Reg_CTPG} for cost gradient estimation. Algorithm \ref{Algo:PO} gives a brief high-level description for the policy optimisation process where the CTPG is adopted for computing the gradients of the cost function with respect to the parameters.

\begin{algorithm2e}[ht!]
	\caption{Policy Optimisation ($\mathsf{PO}$)}
	\label{Algo:PO}
	\DontPrintSemicolon
	\LinesNumbered

	\KwIn{$\mathsf{ODE\_solver}\left(\mathsf{ODE}, \mathsf{IC}, t_{\text{span}}  \right)$, $\mathsf{OPT}\left(\mathbf{p}, J\left(\mathbf{p}\right), \nabla J\left(\mathbf{p}\right)\right)$}
	
	\KwData{$\phi$, $L$, $R$, $\mathbf{f}$, $\mathbf{h}$, $\left(\mathbf{f}_{c},\mathbf{g}\right)\in \mathcal{K}$, $t_{0}$, $t_{f}$, $\mathbf{x}_{0} \in \mathcal{X}_{0}$, $\mathbf{x}_{c_{0}}$, $\mathbf{r}\left(t\right) \in \mathcal{R}$}
	
	\KwOut{$J$, $\frac{dJ}{d\mathbf{p}}$}
	\BlankLine
	$\pi\left(\mathbf{p}\right) \leftarrow$ Construct Neural Network\;
	$\mathbf{p}_{0} \leftarrow$ Initialise$\left(\pi\right)$, \quad $i \leftarrow 0$\;
	\While{$\mathbf{p}_{i}$ is not converged}{
	\ForAll{$\left(\mathbf{x}_{0},\mathbf{r}\left(t\right)\right)_{j} \in \mathcal{X}_{0} \times \mathcal{R}$}{
		$J_{j}\left(\mathbf{p}_{i}\right)$, $\frac{dJ_{j}}{d\mathbf{p}}\left(\mathbf{p}_{i}\right) \leftarrow \mathsf{R}$-$\mathsf{CTPG}\left(\mathbf{p}_{i}; \phi, L, R, \left[\mathbf{f};\mathbf{f}_{c}\left(\mathbf{r}\right)\right], t_{0}, t_{f}, \left[\mathbf{x}_{0}; \mathbf{x}_{c_{0}}\right], \mathsf{ODE\_solver} \right)$\;
	}
	$J\left(\mathbf{p}_{i}\right)$, $\frac{dJ}{d\mathbf{p}}\left(\mathbf{p}_{i}\right) \leftarrow \underset{\forall j}{\mathsf{mean}}\left[ J_{j}\left(\mathbf{p}_{i}\right), \frac{dJ_{j}}{d\mathbf{p}}\left(\mathbf{p}_{i}\right)\right]$\;
	$\mathbf{p}_{i+1} \leftarrow \mathsf{OPT}\left(\mathbf{p}_{i}, J\left(\mathbf{p}_{i}\right), \frac{dJ}{d\mathbf{p}}\left(\mathbf{p}_{i}\right) \right)$, \quad $i \leftarrow i+1$
	}
	\Return{$\mathbf{p}^{*}$, $J\left(\mathbf{p}^{*}\right)$}\;
\end{algorithm2e}

\begin{remark}[Appropriate Adjoint Sensitivity Analysis Technique for Control Problems] \label{Rem:AdjImpl}\\ \noindent 
	Various adjoint sensitivity analysis techniques were compared in \cite{Ainsworth_2021} including the BackPropagation Through Time (BPTT) technique with checkpointing based on Euler discretisation, quadrature adjoint, backsolve adjoint, and interpolating adjoint methods. The backsolve method presented in \cite{Chen_2018} enables memory-efficient adjoint state backpropagation by appending the forward dynamics in the integrand instead of storing full forward solution. However, it is prone to being unstable since the convergent forward pass solution that is usually intended in control problems is divergent in view of reverse mode integration. On the other hand, the BPTT technique is insensitive to the instabilities while performing exact backpropagation, but requires increased computational load to compensate for the loss of accuracy due to discretisation. Hence, the interpolating adjoint method is considered the adequate choice in control problems to perform backpropagation through the ODE model.
\end{remark}

\begin{remark}[Modification of Cost Function with Regularisation Terms] \label{Rem:Reg}\\ \noindent
	The traditional Bolza form cost functional for deterministic optimal control includes the evaluations at discrete time points as well as a continuous functional defined for the Lagrangian with ODE solution. The cost function can also include various regularisation terms; i) a distance function only of NN parameters to avoid overfitting, ii) an information-theoretic function of either parameter or trajectory variables to realise maximum entropy or Bayesian principle \cite{Haarnoja_2018, Kim_2020_maxent, Lambert_2021}, iii) a geometric regularisor of parameters for trust-region penalisation in the context of successive convex programming \cite{Maity_2014,Wagener_2019}, or even iv) a function promoting orthogonality of unstructured expansion basis functions for explainability \cite{Banerjee_2020a}, etc. A possible direction for further investigation is to propose a regularisation term that can reduce the suboptimality gap from the original optimal control problem that does not involve NN parametrisation.
\end{remark}

\section{Example} \label{Sec:NumSim}
This section presents a numerical example to demonstrate the effectiveness of the proposed structured nonlinear controller optimisation framework. The software for both the general purpose training function implementing CTPG and the environment used for the example are available in \cite{Cho_2022}.

\subsection{System Model}
The example considers the normal acceleration tracking control of the nonlinear longitudinal dynamics for a tail-controlled skid-to-turn airframe. The system model can be written as
\begin{equation} \label{Eq:DE_dyn_airframe}
	\begin{aligned}
		\dot{h} &= V\sin\gamma\\
		\dot{V} &= \frac{QS}{m}\left(C_{N}\sin\alpha + C_{A}\cos\alpha \right) - g\sin\gamma\\
		\dot{\alpha} &= \frac{QS}{mV}\left(C_{N}\cos\alpha - C_{A}\sin\alpha\right) + \frac{g}{V}\cos\gamma + q\\
		\dot{q} &= \frac{QSd}{I_{yy}}C_{M}\\
		\dot{\theta} &= q\\
		\ddot{\delta} &= -{\omega_{a}}^{2}\left(\delta-\delta_{c}\right) - 2\zeta_{a}\omega_{a}\dot{\delta}\\
		a_{z} &= -V\dot{\gamma} - g\cos\gamma
	\end{aligned}
\end{equation}
where 
\begin{equation} \label{Eq:DE_dyn_airframe_aux}
	\begin{aligned}
		\gamma &= \theta - \alpha,	\quad Q = \frac{1}{2}\rho V^{2}\\
		\rho  &= \rho_{0}\exp\left(-\frac{h}{H}\right),	\quad V_{s} = \sqrt{\gamma_{a}R_{a}\left(T_{0} - \lambda h\right)}, \quad	M	  = \frac{V}{V_{s}}\\
		C_{A} &= a_{a}\\
		C_{N} &= a_{n}\alpha^{3} + b_{n}\alpha \left|\alpha\right| + c_{n}\left(2-\frac{M}{3}\right)\alpha + d_{n}\delta\\
		C_{M} &= a_{m}\alpha^{3} + b_{m}\alpha \left|\alpha\right| + c_{m}\left(-7+\frac{8M}{3}\right)\alpha + d_{m}\delta
	\end{aligned}
\end{equation}
In Eqs. (\ref{Eq:DE_dyn_airframe})-(\ref{Eq:DE_dyn_airframe_aux}), the variables $h$, $V$, $\alpha$, $\theta$, $\gamma$, $\delta$, $\delta_{c}$ and $a_{z}$ denote the altitude, the speed, the angle-of-attack, the pitch attitude angle, the flight path angle, the control surface deflection, the actuator command, and the normal specific force acceleration which is positive in the nose down direction, respectively. Also, $Q$, $\rho$, $V_{s}$, and $M$ refer to the dynamic pressure, the atmospheric density, the speed of sound, and the Mach number, respectively. The model parameter values borrowed from \cite{Mracek_1997} are summarised in Table \ref{Table:SimParam}.

\begin{table}[ht!] \label{Table:SimParam}
	\begin{center}
		\caption{Simulation Model Parameters}
		\vspace{1em}
		
		\begin{tabular}{cc|cc}
			Quantity & Value & Quantity & Value\\
			\hline\hline
			$a_{a}$			&	$-0.3$					&	$g$			&	$9.8$ [m/s$^{2}$]\\
			$a_{n}$			&	$19.373$				&	$a_{m}$		&	$40.44$\\
			$b_{n}$			&	$-31.023$				&	$b_{m}$		&	$-64.015$\\
			$c_{n}$			&	$-9.717$				&	$c_{m}$		&	$2.922$\\
			$d_{n}$			&	$-1.948$				&	$d_{m}$		&	$-11.803$\\
			$m$				&	$204.02$ [kg]			&	$I_{yy}$	&	$247.439$ [kg$\cdot$m$^{2}$]\\
			$S$				&	$0.0409$ [m$^{2}$]		&	$d$			&	$0.2286$ [m]\\
			$\omega_{a}$	&	$150$ [rad/s]			&	$\zeta_{a}$	& 	$0.7$\\
			$\rho_{0}$		&	$1.225$ [kg/m$^{3}$]	&	$H$			&	$8435$ [m]\\
			$\gamma_{a}$ 	& 	$1.4$ 					&	$R_{a}$		&	$286$ [m$^{2}$/s$^{2}$/K]\\
			$T_{0}$			&	$288.15$ [K]			&	$\lambda$	&	$0.0065$ [K/m]
		\end{tabular}
	\end{center}
\end{table}

The controller structure can be prescribed as the Raytheon three-loop autopilot which can be expressed as
\begin{equation} \label{Eq:3LA}
	\begin{aligned}
		\dot{x}_{c} &= K_{A}\left(a_{z_{cmd}}-a_{z}\right) + q + \frac{a_{z_{cmd}} +g\cos\gamma}{V}\\
		\delta_{c} 	&= K_{I}x_{c} + K_{R}q
	\end{aligned}
\end{equation}
where $a_{z_{cmd}}$ denotes the acceleration command with $K_A$, $K_{I}$, and $K_{R}$ being the tunable elements. 

\subsection{Neural Networks Training}
Let us consider the NN parametrisation given by
\begin{equation} \label{Eq:NN_parametrisation}
\begin{bmatrix}
	K_{A}\\
	K_{I}\\
	K_{R}
\end{bmatrix}
= \pi\left(\frac{\left|\alpha\right|}{\alpha_{\max}}, \frac{M}{M_{\max}}, \frac{h}{h_{\max}}, \mathbf{p}\right)
\end{equation}
where the subscript $\max$ denotes the characteristic maximum value for each variable introduced for normalisation of the NN input. Then, Eq. (\ref{Eq:3LA}) becomes a neural feedback controller. As explained previously, the training objective for structured controller optimisation is in background learning of the state-dependent policy rather than the decision-time planning of control inputs. Therefore, the NN training is performed by leveraging ensemble simulation for a batch of initial altitude $h_{0}$, initial speed $V_{0}$, and acceleration command $a_{z_{cmd}}$. Note that random sampling of the initial state and the command is another possibility for ensemble construction.

The control-theoretic knowledge about the autopilot structure suggests to limit the possible range for each NN output to some interval $\left(K_{lb}, K_{ub}\right)$ to ensure closed-loop stability by prescribing the right sign for each gain and to avoid instabilities due to excessive gains. In this example, the NN output $K_{A}$, $K_{I}$, and $K_{R}$ should take positive values for stable command tracking. This prior knowledge can be incorporated into the NN construction by placing a scaling layer at the output end for which the node activation rule is given by $K_{lb} + \left(K_{ub}-K_{lb}\right)\sigma\left(x\right)$ where $\sigma\left(x\right)$ is a sigmoid function that takes its value in $\left(0,1\right)$. Note that the scaling operation applies to a vector in an elementwise manner.

In practice, the most challenging part of the policy optimisation process is the choice of a good cost function representing the design objectives. In the feedback controller parameter optimisation problem, the difficulties arise from the necessity to meet multiple design criteria associated with the transient response out of a single scalar cost function. The demand to keep a consistent response shape across a wide range of initial conditions and commands introduces additional complexity. As discussed earlier in \cite{Shin_2020}, the discrepancy between the actual response and a reference model for the closed-loop command tracking can encode the response shaping necessities in the cost function. For this purpose, this study considers the following reference model
\begin{equation} \label{Eq:a_z_ref}
	\dot{a}_{z_{ref}} = \frac{a_{z_{cmd}} - a_{z_{ref}}}{0.2}
\end{equation}
where $a_{z_{ref}}$ represents the reference acceleration. For the training scenario consisting of multiple command values, the contribution of each case to the cost function defined by the ensemble mean as decribed in Eq. (\ref{Eq:Prob_StructOC}) should be comparable to each other to promote consistency of transient response. For this reason, the associated running cost term is defined by $\left|\frac{a_{z}-a_{z_{ref}}}{1+\left|a_{z_{cmd}}\right|}\right|^{2}$ considering normalisation with respect to the command instead of a fixed maximum value. Here, the addition of $1$ in the denominator is to permit the case of $a_{z_{cmd}}=0$.

The training scenario, the NN layer configuration, and the optimiser hyperparameters are summarised in Table \ref{Table:NNTrainConfig}. Entire computation is performed with CPU multi-threading on a laptop equpped with a 2.8GHz quad-core Intel Core i7 CPU and 16GB 2133MHz LRDDR3 RAM.

\begin{table}[hbt!] \label{Table:NNTrainConfig}
	\begin{center}
		\caption{Training Setup and Neural Networks Configuration}
		\vspace{1em}

		\resizebox{0.85\textwidth}{!}{
		\begin{tabular}{cccc}
			Object		& Component 			&	Description\\
			\hline\hline
\multirow{3}{*}{$\mathbf{x}_{0}$} & $\alpha_{0}$, $q_{0}$, $\theta_{0}$, $\delta_{0}$, $\dot{\delta}_{0}$ &	$0$\\
						& $h_{0}$				&	$5000:1000:8000$ [m]\\
						& $V_{0}$				&	$700:100:900$ [m/s]\\
		$\mathbf{r}$	& $a_{z_{cmd}}$			&	$-100:25:100$ [m/s$^{2}$]\\
			\hline		
\multirow{3}{*}{$J$} 	& $L$					&	$100\left|\frac{a_{z}-a_{z_{ref}}}{1+\left|a_{z_{cmd}}\right|}\right|^{2} + 0.01\left|\frac{\delta_{c}}{5\pi/36}\right|^{2} + 0.1\left|\frac{\dot{\delta}}{1.5}\right|^{2}$\\
						& $\phi$				&	$0$\\
						& $R$					& 	$10^{-4}\left\|\mathbf{p}\right\|_{2}^{2}$\\
			\hline
\multirow{5}{*}{$\pi$} 	&  input layer			&	3 $\tanh$\\
						& hidden layer			&	10 $\tanh$\\
						& output layer			&	3 linear\\
						& scaling layer			&	$K_{lb} + \left(K_{ub} - K_{lb}\right)\sigma\left(x\right)$\\
						&$\left(\alpha_{\max}, M_{\max}, h_{\max}\right)$	& $\left(\pi/6, 4, 11\,000\right)$\\
						& $\left(K_{lb}, K_{ub}\right)$ & $\left(\begin{bmatrix}10^{-3} & 10^{-3} & 10^{-3}\end{bmatrix}^{T}, \begin{bmatrix}4 & 0.2 & 2\end{bmatrix}^{T}\right)$\\
	\hline
	\multirow{5}{*}{ODE solver} 	&	solve algorithm	&	\textsf{Tsi5} (Tsitouras 5/4 Runge-Kutta method)\\
						&	integration time span	&	$\left[0,3\right]$ [s]\\
						&	solution saving step & $10^{-2}$ [s]\\
						&	absolute tolerance	&	$10^{-6}$\\
						& 	relative tolerance	&	$10^{-3}$\\
			\hline
	\multirow{7}{*}{optimiser} & \multirow{4}{*}{1st phase} & \textsf{ADAM}\\
						&						&	learning rate $\eta= 0.01$\\
						&						& momentum decay $\beta=\left(0.9,0.999\right)$\\
						&	& maximum iteration $=1\,000$\\
	 & \multirow{3}{*}{2nd phase} & \textsf{BFGS}\\
						&						&	initial step norm $= 10^{-4}$\\
						&						& maximum iteration $=1\,000$
		\end{tabular}
		}
	\end{center}
\end{table}

\subsection{Results and Discussions}
\subsubsection{Learning Efficiency}
The numerical experiment compares three different setups described in Table \ref{Table:Cases} to illustrate the strengths of continuous-time method for computing policy gradients and the importance of structural knowledge in NN optimisation. 

\begin{table}[hbt!] \label{Table:Cases}
	\begin{center}
		\caption{Experiment Cases}
		\vspace{1em}

		\begin{tabular}{lc}
			Case Title		& Description\\
			\hline\hline
			\texttt{base}			& baseline setup described in Table \ref{Table:NNTrainConfig}\\
			\texttt{unscaled}		& scaling layer in NN is removed from baseline\\
			\texttt{discrete} 		& Euler method with fixed step size of $10^{-3}$[s] is used for integration
		\end{tabular}
	\end{center}
\end{table}

\begin{figure}[hbt!]
	\begin{center}
		\includegraphics[width=0.85\textwidth]{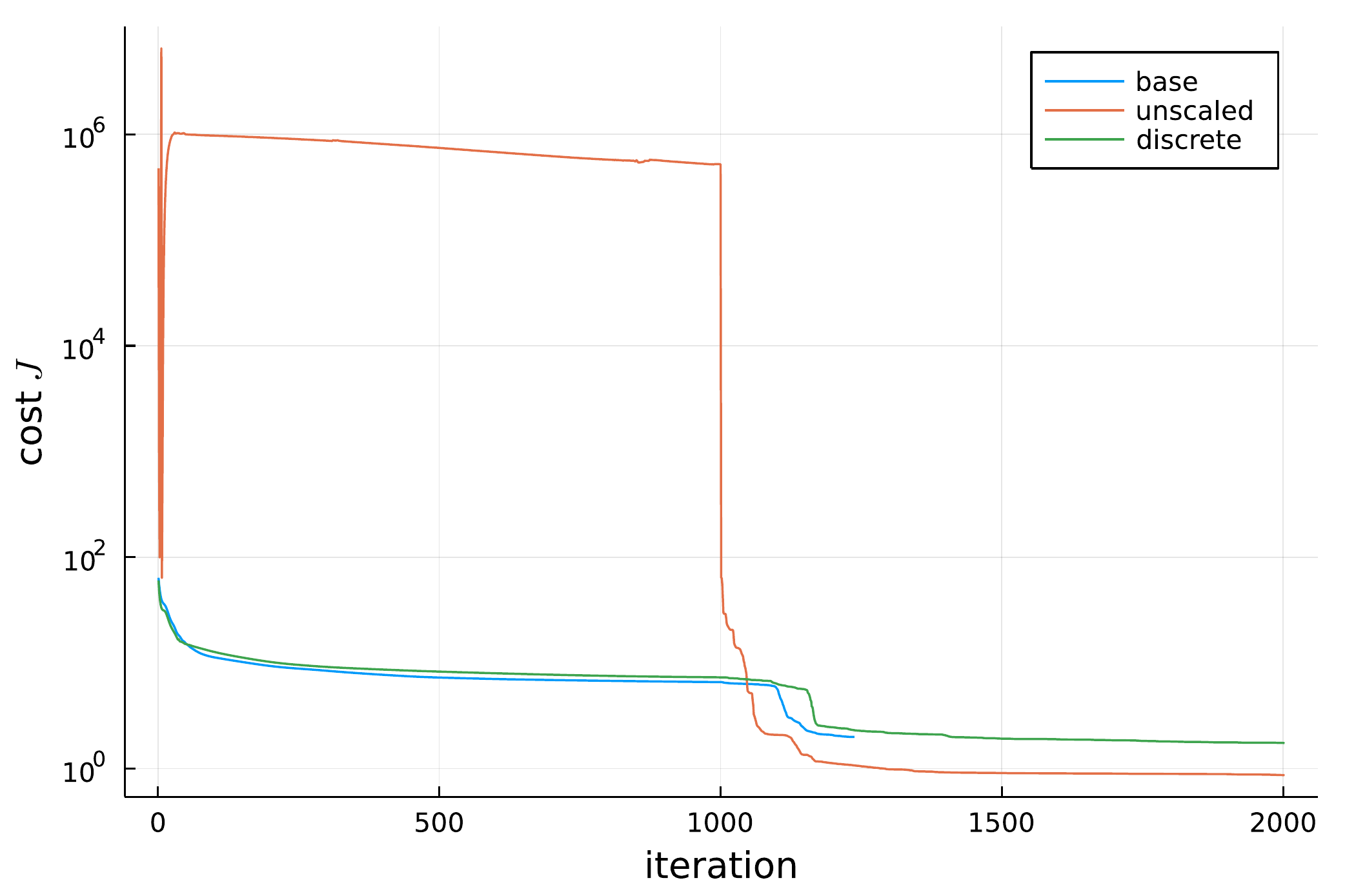}
		\caption{Cost History (Learning Curve)}
		\label{Fig_J}
	\end{center}
\end{figure}

Table \ref{Table:Results} summarises the results of NN policy learning for each case. The comparison clearly indicates that the learning efficiency substantially depends on the gradient computation method as well as the NN structure. Overall, the result implies that the main benefit of the CTPG method is the rate of convergence and the computation time required for convergence. The \texttt{base} case converged to a local minimum with the least amount of optimisation iterations and wall-clock time while resulting in an optimal cost comparable to the \texttt{discrete} case at the end of learning. The learning curves shown in Fig. \ref{Fig_J} also supports the same argument. On the other hand, the \texttt{unscaled} case suffered from the divergence of numerical integration that is frequently observed during the initial phase before the parameters being located at a point rendering the closed-loop system stable. The value of the final cost obtained for \texttt{unscaled} case is considered unrealistic since the converged result is obtained with excessive values for the gains that lacks robustness. The \texttt{discrete} case showed the fragility of the discretised backpropagation through time as it was prone to complete failure during the learning process due to numerical instability. A small step size no greater than $10^{-3}$ [s] was necessary for the solver stability to avoid catastrophic optimisation failure, however, the reduction of the step size inevitably leads to the increased computational cost. Also, the policy learning process was exited by reaching the given maximum number of iterations without convergence.

\begin{table}[hbt!] \label{Table:Results}
	\begin{center}
		\caption{Summary of Policy Learning Results}
		\vspace{1em}
		
		\begin{tabular}{cccc}
									& \texttt{base} 		& \texttt{unscaled} & \texttt{discrete}\\
			\hline\hline
			final cost $J^{*}$		& $1.993292$ & 	$0.866841$			&	$1.744865$\\
			number of iterations	& $1238$	&	$2002^{\dagger}$	&	$2002^{\dagger}$\\
			computation time [s]	& $5888.4$	& $18419.3$	&	$20293.3$\\
			average computation time per iteration [s] & $4.756$ & $9.200$ & $10.137$\\
			training failure 		& none 		& frequent divergence & frequent failure
		\end{tabular}
		\\[0.5em]
		{\footnotesize $\dagger$: terminated by reaching maximum number of iterations}
	\end{center}
\end{table}

\subsubsection{Simulation with Optimised Policy}
\begin{figure}[ht!]
	\begin{center}
		\includegraphics[width=0.85\textwidth]{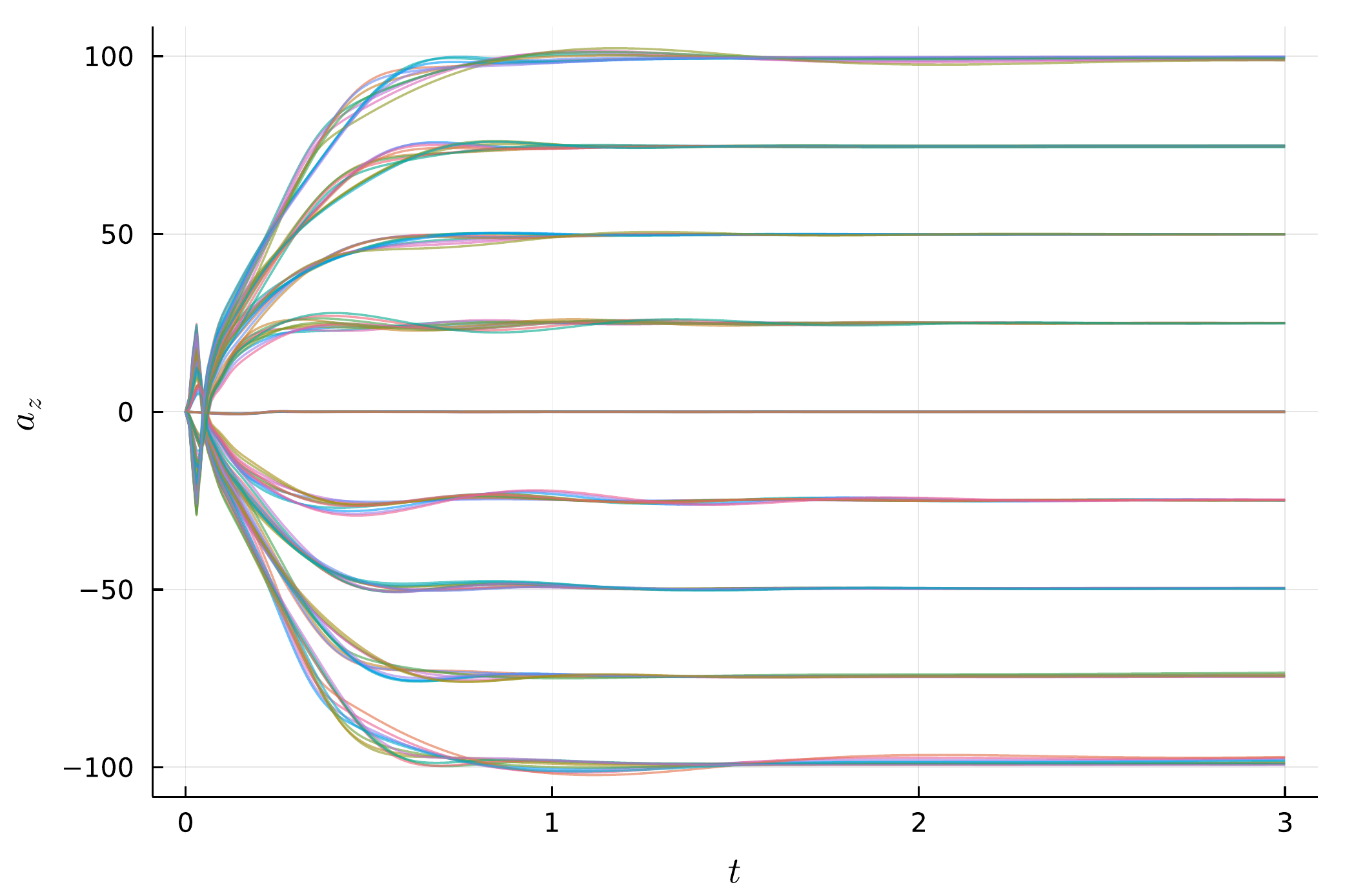}
		\caption{Output History}
		\label{Fig_y}
	\end{center}
\end{figure}

\begin{figure}[ht!]
	\begin{center}
		\includegraphics[width=0.85\textwidth]{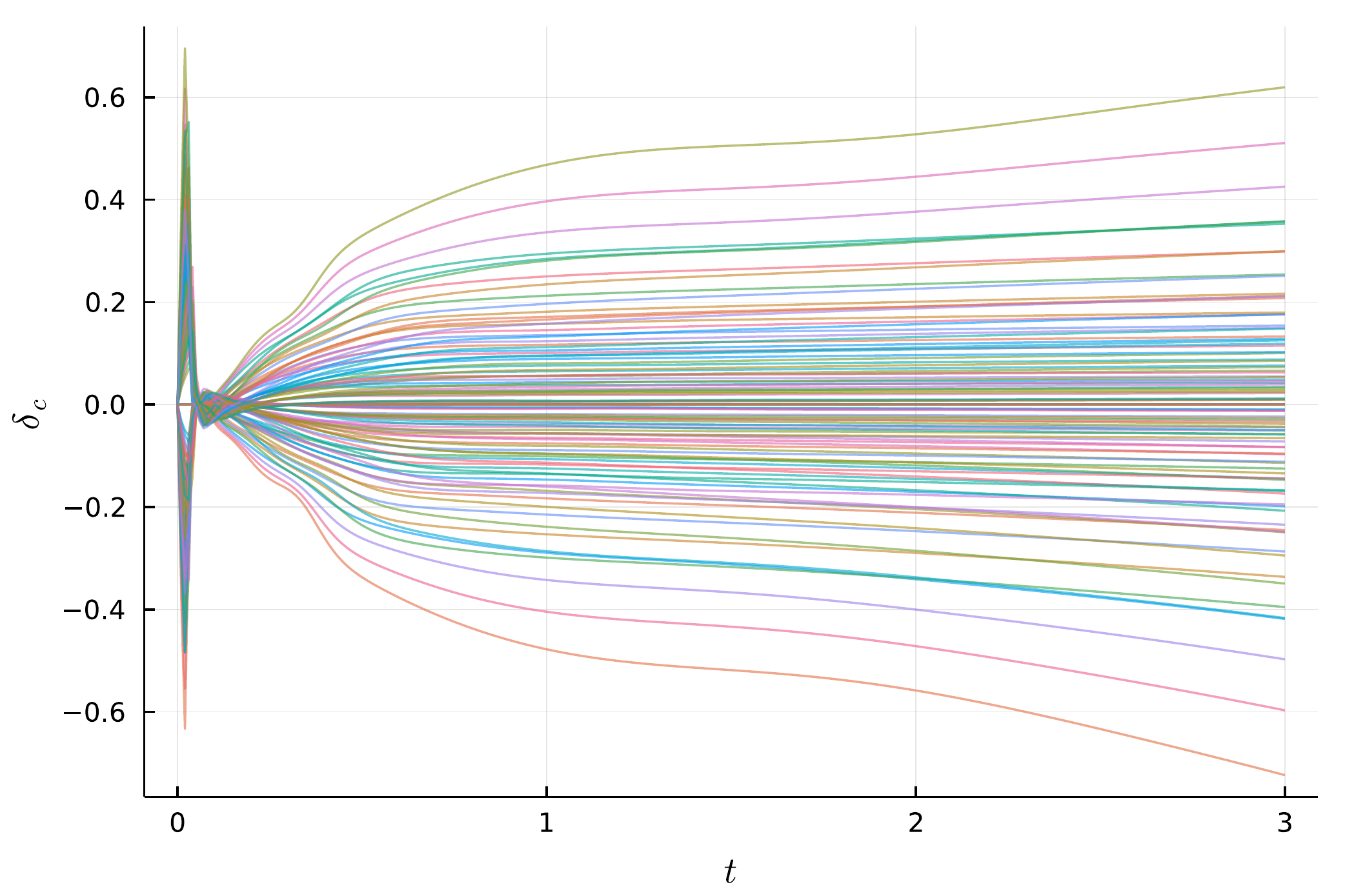}
		\caption{Input History}
		\label{Fig_u}
	\end{center}
\end{figure}

\begin{figure}[ht!]
	\begin{center}
		\includegraphics[width=0.85\textwidth]{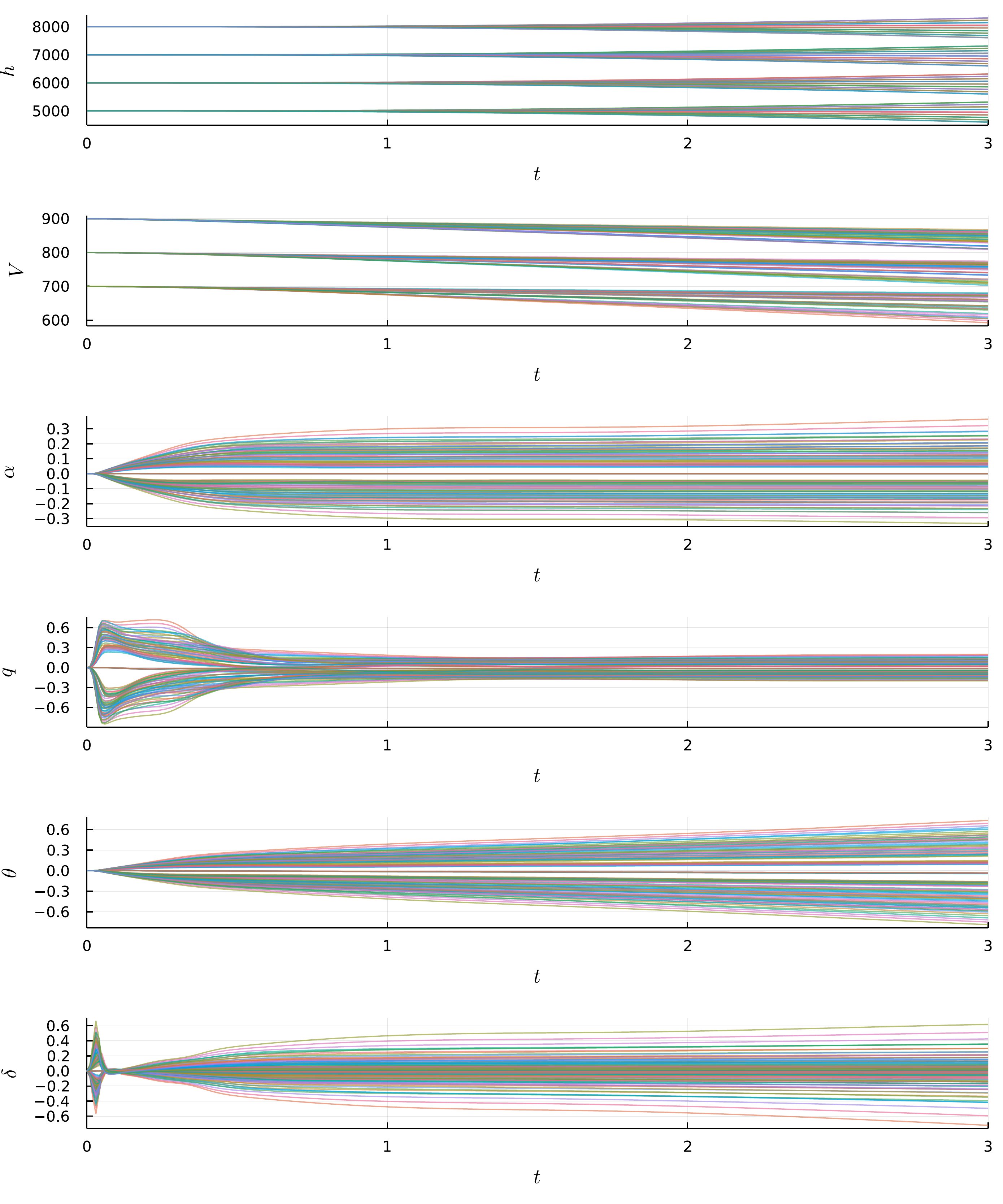}
		\caption{State History}
		\label{Fig_x}
	\end{center}
\end{figure}

This section presents the simulated response of the optimised controller for the \texttt{base} case. Figures \ref{Fig_y}-\ref{Fig_x} show the time histories of the output, the state, and the input, respectively. The learned policy produces consistent transient responses over various initial states and commands as shown in Figs. \ref{Fig_y} and \ref{Fig_x}. The learned policy achieves successful tracking of the commanded normal acceleration.

\begin{figure}[ht!]
	\begin{center}
		\includegraphics[width=0.8\textwidth]{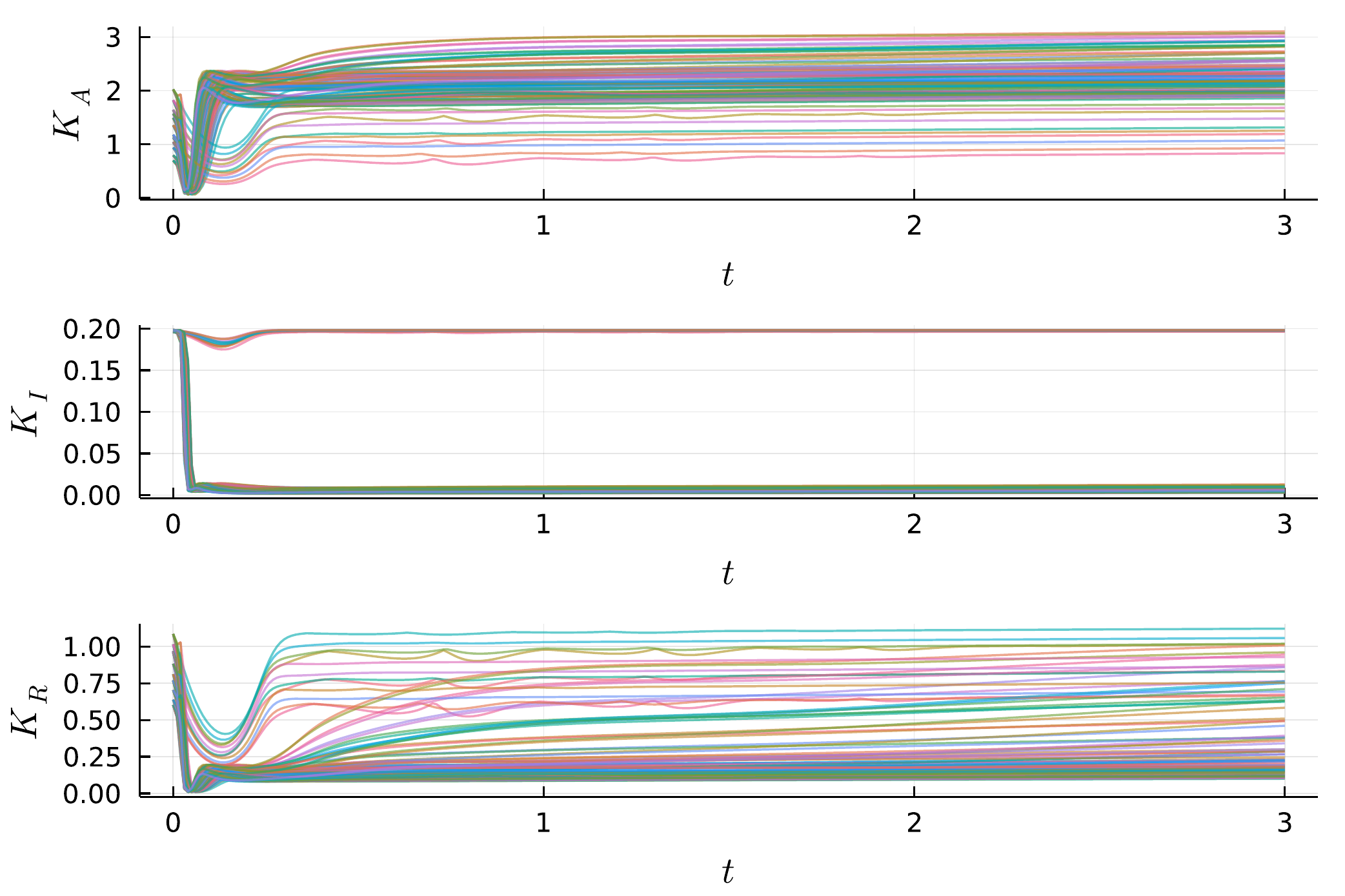}
		\caption{Gain History}
		\label{Fig_y_NN}
	\end{center}
\end{figure}

\begin{figure}[ht!]
	\begin{center}
		\subfigure[$K_{A}$]{
			\includegraphics[width=0.4\textwidth]{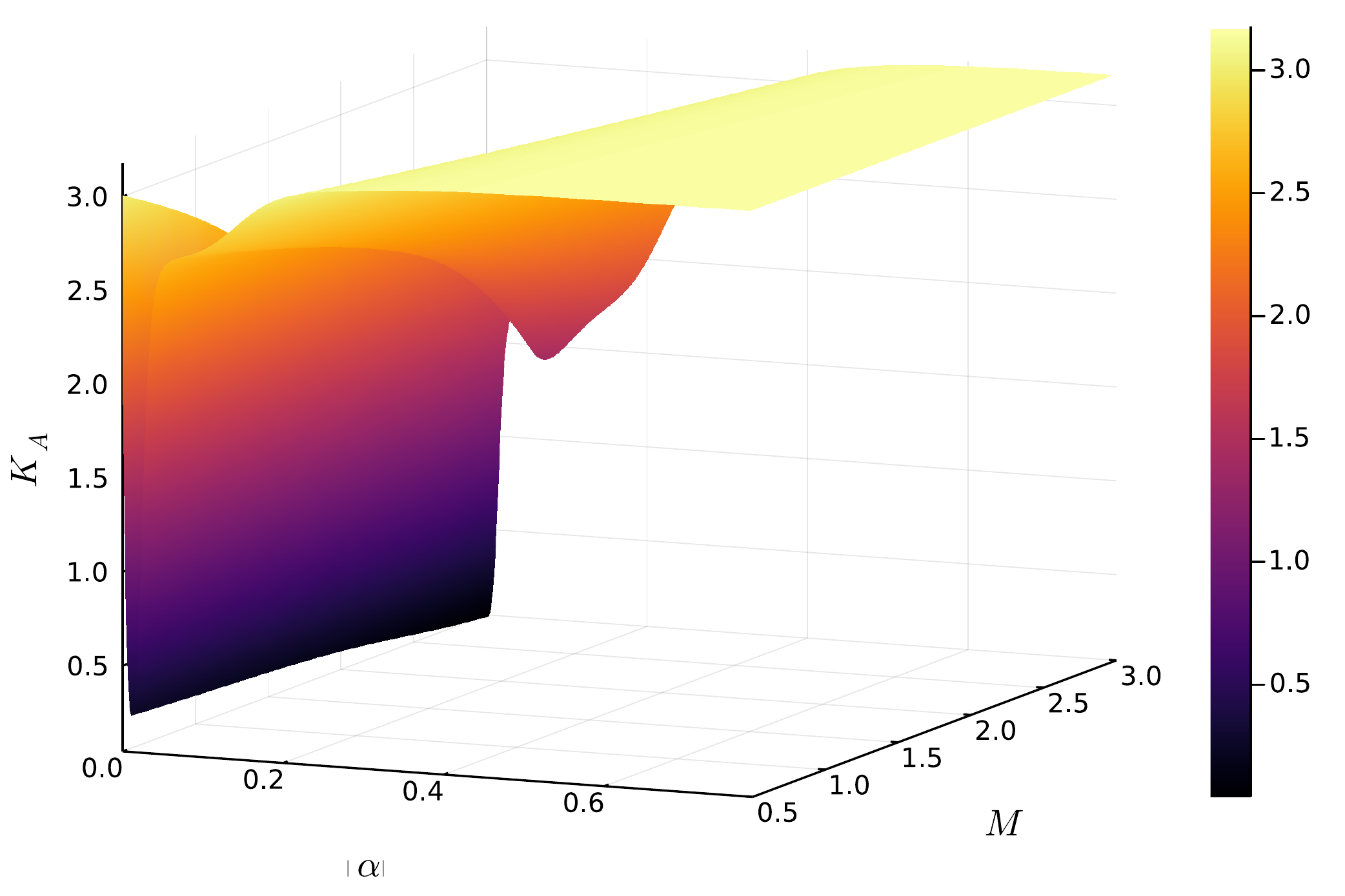}
		}
		~
		\subfigure[$K_{I}$]{
			\includegraphics[width=0.4\textwidth]{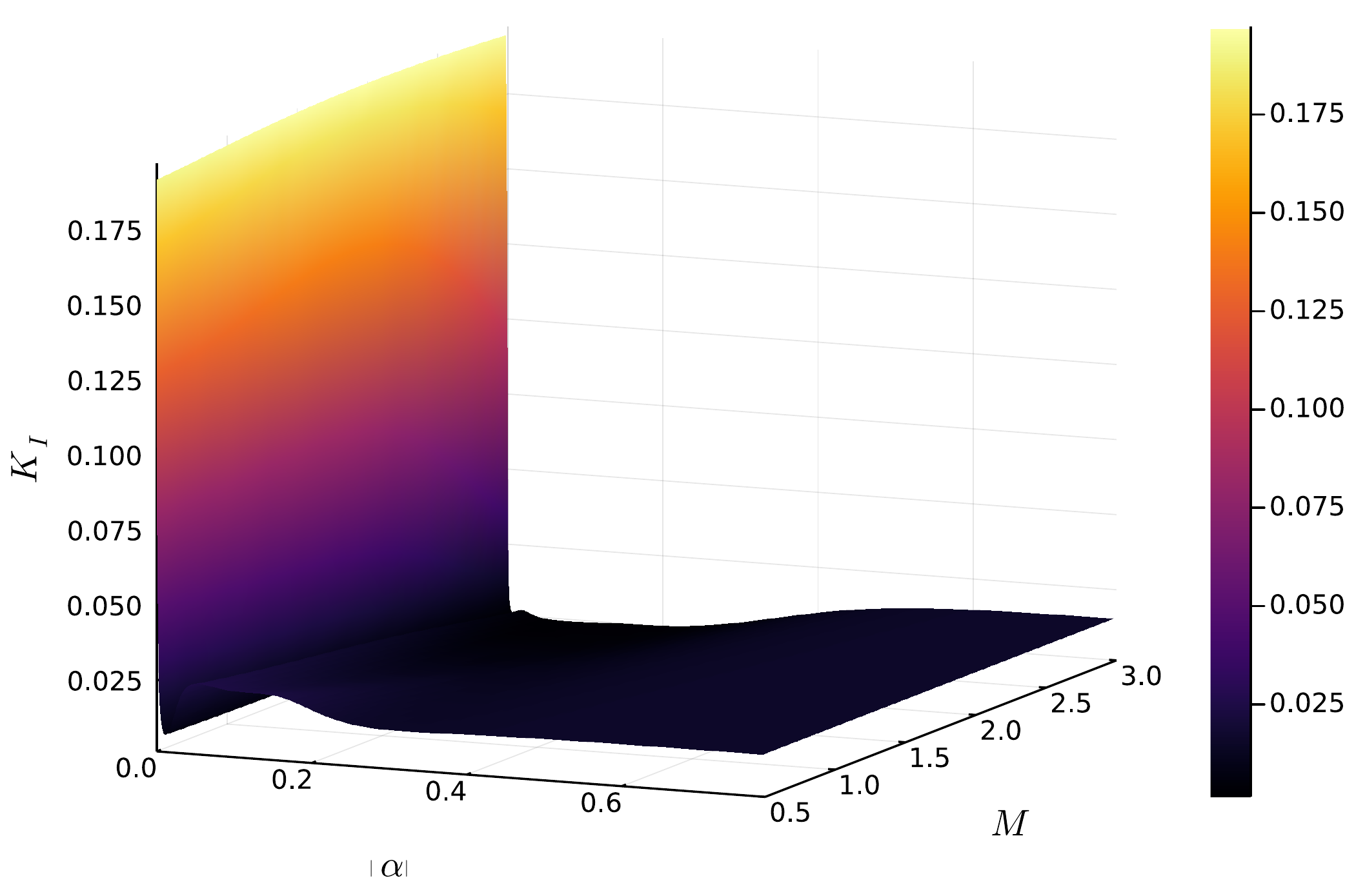}	
		}
		~
		\subfigure[$K_{R}$]{
			\includegraphics[width=0.4\textwidth]{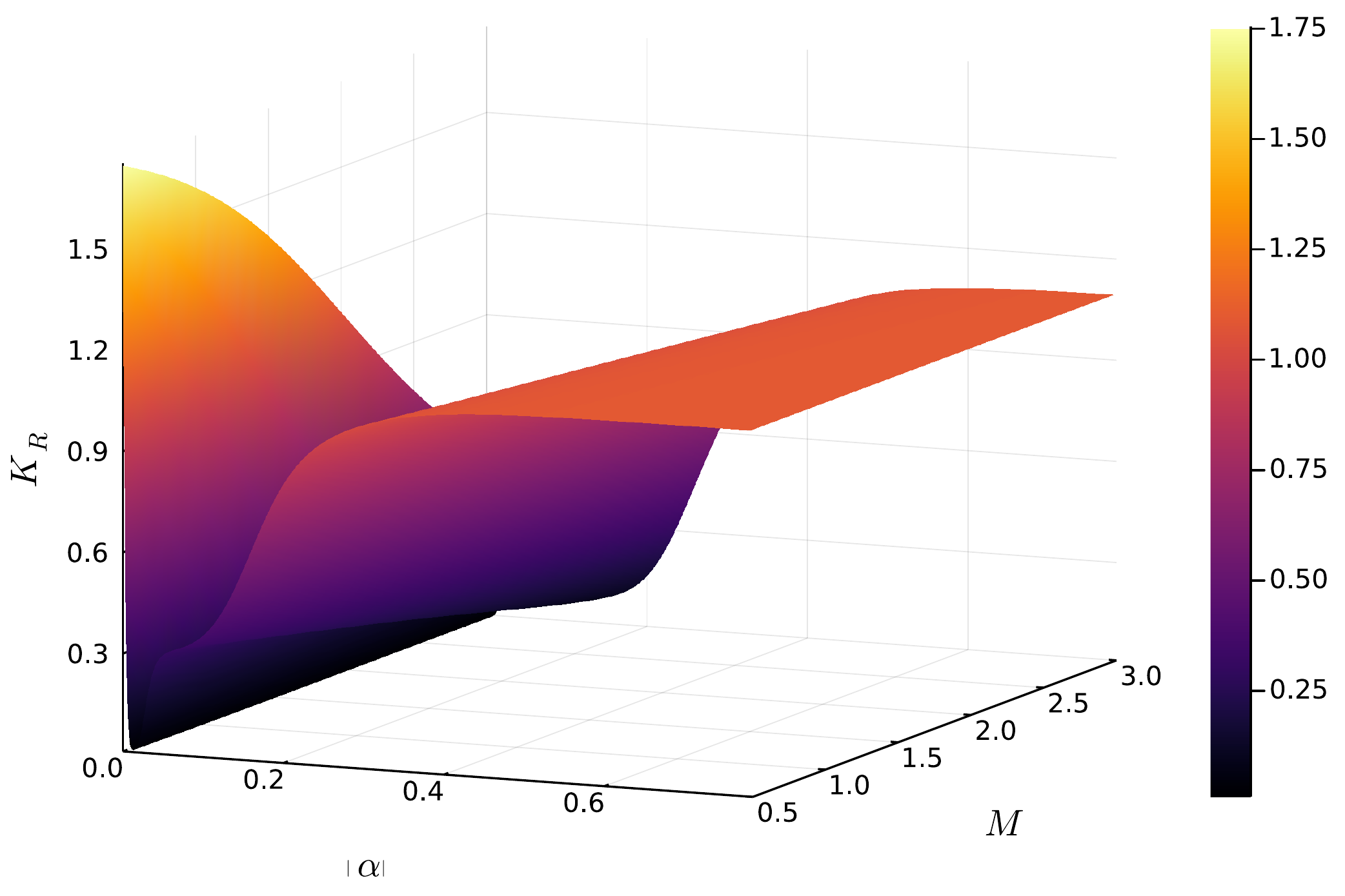}
		}
	\end{center}
	\caption{Gain Surfaces Evaluated at $h=5\,000$ [m]}
	\label{Fig_K}
\end{figure}

Figure \ref{Fig_y_NN} shows the NN output, i.e., controller gains, as a time history, and Fig. \ref{Fig_K} shows the learned policy evaluated at $h= 5\,000$ [m]. One benefit of using NN parametrisation for the policy is that the resulting gains are smooth functions which are optimised directly through nonlinear time-domain simulation. This approach can save the efforts required in the post-processing of gain schedules given by look-up tables, which is common in practical gain scheduling design process based on linear synthesis techniques.

The result of policy learning depends on the problem formulation considered for policy optimisation. From our experiments, a higher weight given to the regularisor $R$ resulted in a more moderate change in the gains with respect to time. However, too large weighting for the $\mathcal{L}_{2}$ regularisation governs the trend of the overall cost landscape and hence the optimal solution. Also, a wrong setting of the gain upper bound $K_{ub}$ for the scaling layer can produce unstable initialisation that deteriorates the learning efficiency. Furthermore, the time-domain response characteristics and the gain histories exhibited dependence on the number of nodes in the hidden layer. Choosing a large number of hidden layer nodes easily resulted in overfitting of the NN parameters so that the gains attain values close to the bounds specified by the scaling layer.

\section{Conclusions} \label{Sec:Concls}
A method for computing the gradient of the cost functional with respect to the policy parameters was studied considering the continuous-time nature of the system dynamics. The continuous-time policy gradient (CTPG) method is essentially based on the adjoint sensitivity analysis techniques for studying ordinary differential equations (ODEs). Specifically, this study addressed application of the CTPG method for performance optimisation of structured control systems prescribing a well-known structure for the actual control input variable while considering tunable parameters as the parametrised policy, focusing on combining the benefits of existing scientific knowledge with the capabilities of machine learning. 

Numerical experiments considering optimisation of the gains in three-loop acceleration autopilot for a flying vehicle model demonstrated that the CTPG method yields higher learning efficiency and better optimality of the final outcome in comparison to the backpropagation based on explicit time-discretisation of the system dynamics. The results suggest that the use of available ODE solvers capable of adaptive time-stepping enables correct and efficient computation of the gradient with the help of software tools developed for differentiable programming. Also, the tradeoff between computational efficiency and solution accuracy can be systematically adjusted through solver tolerance control. In these regards, the CTPG method holds potentials to improve the effectiveness of direct neural policy optimisation in many continuous-time optimal control / reinforcement learning tasks. 

\clearpage
\bibliography{CTPG_struct_ctrl}
\end{document}